\documentclass[runningheads]{llncs}
\usepackage[T1]{fontenc}
\usepackage{graphicx}
\usepackage{float}
\begin{document}
\title{Optical Character Recognition using Convolutional Neural Networks for Ashokan Brahmi Inscriptions}
\titlerunning{OCR using CNN for Ashokan Brahmi Inscriptions}
%
\author{Yash Agrawal \and Srinidhi Balasubramanian\and Rahul Meena \and Rohail Alam \and Himanshu Malviya \and Rohini P}
%
\authorrunning{Y. Agrawal et al.}
%
\institute{IIITDM Kancheepuram, Chennai, India}
\maketitle              
\begin{abstract}
This research paper delves into the development of an Optical Character Recognition (OCR) system for the recognition of Ashokan Brahmi characters using Convolutional Neural Networks. It utilizes a comprehensive dataset of character images to train the models, along with data augmentation techniques to optimize the training process. Furthermore, the paper incorporates image preprocessing to remove noise, as well as image segmentation to facilitate line and character segmentation.

The study mainly focuses on three pre-trained CNNs, namely LeNet, VGG-16, and MobileNet and compares their accuracy. Transfer learning was employed to adapt the pre-trained models to the Ashokan Brahmi character dataset. The findings reveal that MobileNet outperforms the other two models in terms of accuracy, achieving a validation accuracy of 95.94\% and validation loss of 0.129. The paper provides an in-depth analysis of the implementation process using MobileNet and discusses the implications of the findings.

The use of OCR for character recognition is of significant importance in the field of epigraphy, specifically for the preservation and digitization of ancient scripts. The results of this research paper demonstrate the effectiveness of using pre-trained CNNs for the recognition of Ashokan Brahmi characters.

\end{abstract}
\section{Introduction}

Brahmi is an ancient pan-Indian national script, that emerged as a fully developed script in the 3rd century BCE and is considered the parent of many modern Indic scripts. Scholars differentiate the Brahmi of different eras - ‘Early’, ‘middle’, and ‘late’. In the Mauryan period (3rd century BCE), we find Early Brahmi used extensively in the Ashokan rock, pillar, and cave inscriptions. With some regional variations, it was largely uniform and almost fully developed script. Starting from the 3rd century, Brahmi and its  derivatives continued to be in use until the 4th century CE. As Brahmi continued to evolve in the modern regional scripts, the ability to write and read old Brahmi was forgotten and was rediscovered in the 19th century \cite{salomon1998indian}. Brahmi has traditionally been an area of study only for linguists and epigraphists, but as the volume and complexity of Epigraphy has increased exponentially, it is now imperative to adapt state-of-the-art technologies to aid the preservation and study of Epigraphy. Optical Character Recognition (OCR) can make previously inaccessible epigraphical data available by converting them into digital, machine-readable data. 

While the first OCR technology was introduced nearly a century ago, it is only in this decade that saw a tremendous transformation in the performance and adaptability of OCR in a wide variety of applications and this was a direct result of using advanced image processing algorithms and deep learning models. Today, several off-the-shelf AI models are used for OCR with many of them exhibiting near human-level performance in specific use cases. However, the majority of these models have been developed for modern popular scripts, with little or no progress in development of OCR models for historical or extinct scripts. One of the greatest hindrance in research and development of such OCR models is the lack of large quantities of data required to train sophisticated deep learning models.

Transfer learning significantly alleviates the challenges posed by small datasets, as they allow us to fine-tune models pre-trained on large datasets using a relatively limited dataset of Brahmi characters. This paper explores the use of three pre-trained CNNs - LeNet, VGG-16 and MobileNet by transfer learning characters of Ashokan Brahmi inscriptions.

\section{Literature Survey}

The development of OCR systems for epigraphy has been an area of growing interest in the recent years. This literature review examines several key research works that have contributed to the field of OCR for inscriptions, and more specifically on Brahmi. While earlier research focused on rule-based algorithms for print characters, this literature survey will focus on recent methods using neural networks, focusing on their dataset, methodologies and results

In 2015, Sowmya and Kumar \cite{fist_lit} demonstrated an Artificial Neural Network (ANN) classifier by extracting Gabor and Zonal features. A dataset of 1032 Ashokan Brahmi characters across 258 classes were used, with a resultant accuracy of   80.2\%. 

In 2017, a study by Manigandan et al. \cite{8389589} proposed recognition of various Tamil characters between the 9th and 12th centuries using OCR and NLP techniques. The authors used pre-processed and segmented images of inscriptions. First, the colored images were converted into gray and then to binary based on a threshold value, then features were extracted using SIFT algorithms for each letter, then characters were classified and constructed using an SVM classifier, and finally, the patterns were matched with known characters and predicted using the Trigram method. A dataset of 40 images from each century was used for training and 10 for testing. Its accuracy varied on the quality of the input image.

Another study in 2017 by Gautam and Chai \cite{2017survey}, focused solely on Brahmi character recognition. A key component of their approach is the geometric method used for feature extraction, which categorizes Brahmi characters into six distinct entities. These features are then used for classification. However, this study was done on printed and handwritten characters instead of epigraphic images, with 
a maximum accuracy of 94.90\% and 91.69\% respectively.

In 2019, a study by Wijerathna et al. \cite{9103340} proposed a CNN-based recognition and translation system for Ashokan Brahmi characters into Sinhala. The authors used image processing techniques to identify correct letters and the time period of inscription using CNN in deep learning. The letters were translated into modern Sinhala letters using NLP. A dataset of 1500 images of Ashokan Brahmi characters was used for training and 300 for testing. An accuracy of 93.33\% on the test set using VGG16 was achieved.

\section{Methodology}

\subsection{Dataset Creation}
The dataset used in this study consists of images of Ashokan Brahmi characters. It was manually created using images from the Indoskript - Manuscripts project, which is a paleographic database of Brahmi and Kharosthi that aims at covering the full historical development of these Indic scripts. Indoskript can be queried for different criteria such as historical time, place of origin, or phonetic restrictions.

The initial dataset consisted of a total of 3500 good-resolution, less-noise images of the Ashokan Brahmi characters. It included 6 vowels and 208 consonants, totaling 214 unique characters.

\subsection{Dataset Augmentation}
CNN was used to classify images of Ashokan Brahmi characters. A large dataset of images was needed to train the CNN effectively. To create this dataset, the data augmentation technique was employed. Various transformations were applied to the images in the dataset, such as rotating, scaling, shifting, and shearing. The contrast and brightness levels of the images were also changed \cite{Data_Augmentation}.

The angle of rotation was randomly chosen between 0 and 10 degrees in both the clockwise and counter-clockwise directions. The images were scaled by a factor of 0.2, which involves either increasing or decreasing the size of the image. The images were also height and width shifted by a factor of 0.15, which involves moving the images along the vertical or horizontal axes. All these transformations were used in combination to expand the amount of training data, making the model more robust and less prone to over-fitting.

As a result of this process, each character in the dataset had roughly 1000 modified images. In total, close to 227000 images were created for the 214 characters in the dataset. This represented a significant increase from the original dataset and allowed for more effective training of the CNN.

\subsection{Image Preprocessing}

Inscription images are noisy, with limited contrast between the letters (foreground) and the background. To address this, median blur was implemented to remove noise, followed by Otsu thresholding for effective foreground-background classification.

The Salt-and-Pepper noise present in the images was removed using median blur, a nonlinear digital filtering technique that replaces each pixel in an image, preferably a grayscale image, with the median value of neighbouring pixels within a fixed kernel size.

After applying median blur, the images were thresholded using Otsu thresholding. This method is preferred over other techniques, such as simple thresholding and adaptive thresholding, due to its ability to select the optimal threshold value. The algorithm divides an image into two distinct classes: class C1, containing possible intensities from 0 to T, and class C2, containing possible intensities from T+1 to 255, where T is the threshold value. One of the 2 classes is designated as the foreground and the other as the background. The optimal threshold value is selected to minimize the within-class variance, ensuring the best possible separation between the foreground and background.

\subsection{Image Segmentation}
Inscription images typically contain several lines, with each line consisting of multiple words. However, the distinction between words is not always clearly evident, as the spaces between words and characters are uniform. Therefore, it is necessary to segment an inscription image into individual lines and further into individual characters, which can then be processed by a classification neural network. This segmentation was achieved using projection profile methods, a widely employed technique that has been specifically explored for Brahmi inscription images \cite{segmentation}.

The horizontal projection profile method takes advantage of the absence of black pixels in the gaps between lines to separate a paragraph into individual lines. The vertical projection profile method uses the same logic vertically to separate different characters. 

\subsection{Classification Models}
Convolutional Neural Networks (CNNs) are a class of Deep Learning (DL) models  that are extensively utilized for image recognition and classification tasks due to their ability to automatically and adaptively learn spatial hierarchies of features through backpropagation. It leverages local connections through convolutional layers, pooling layers, and fully connected layers, to capture spatial hierarchies and patterns effectively.  The hierarchical architecture of CNNs enables the model to capture complex patterns and features within images, making them highly effective for OCR tasks \cite{CNN}.

Transfer learning is technique where a pre-trained model developed for a specific task is repurposed as the foundational model for a different but related task. This approach is particularly beneficial in deep learning domains such as computer vision and natural language processing, where training the models from scratch is computationally intensive and requires vast amounts of data. By leveraging the learned features from large-scale datasets, transfer learning accelerates the training process, enhances model performance, and mitigates the risk of overfitting when dealing with limited domain-specific data \cite{Transfer_Learning}.

In this study, we employ transfer learning using three pre-trained models - LeNet, VGG-16, and MobileNet for performing OCR on Ashokan Brahmi inscriptions. 

\subsubsection{LeNet} : The LeNet model, consisting of five layers with trainable parameters, employs two convolutional layers combined with Max Pooling for spatial feature extraction. The convolution layers in LeNet used the sigmoid activation function, which is less computationally expensive and beneficial for simpler architectures.

Following the convolutional and Max Pooling layers in the LeNet model, two fully connected layers are employed. To feed data into these fully connected layers, each sample is flattened into a two-dimensional array using a flattening layer. The model comprises 323,826 trainable parameters and was trained for 12 epochs, with early stopping based on validation loss, allowing for a patience of 5 epochs \cite{LeNet}.

\subsubsection{VGG-16}: The VGG-16 model, is a transfer learning model pre-trained on the ImageNet dataset. It features 16 layers with weights, including five Max Pooling layers and three Dense layers, totaling 21 layers. The convolutional layers in VGG-16 also utilizes the sigmoid activation function, while the fully connected layers employ the Exponential Linear Unit (ELU) activation function \cite{VGG16}. The model contains 324,414 trainable parameters and was trained for 17 epochs using an early stopping metric based on validation loss, measured by sparse categorical cross-entropy.

\subsubsection{MobileNet}: 
MobileNet is a class of efficient CNN models designed specifically for resource-constrained environments, such as mobile and embedded devices. The efficiency of MobileNet stems from its use of depthwise separable convolutions, which decompose a standard convolution into a depthwise convolution and a pointwise convolution. This decomposition significantly reduces the number of parameters and computational complexity without substantial loss in accuracy, enabling real-time inference on devices with limited processing power. \cite{MobileNet}.

The neural network architecture included a single hidden layer with 128 nodes, utilizing the ELU activation function to enable faster convergence and better learning. The target variable is predicted using the softmax activation function, which gives probabilistic predictions across multiple classes. Figure 1 shows the architecture of MobileNet Model used based on depth-wise separable filters.

For the MobileNet model, two pooling techniques were explored: Max pooling and average pooling. Average pooling was trained for 49 epochs with the early stopping callback, halting if validation loss does not improve for 4 consecutive epochs. Max pooling was trained for 22 epochs using the same early stopping metric.

\begin{figure}[!]
  \begin{center}
  \includegraphics[width=2.9in]{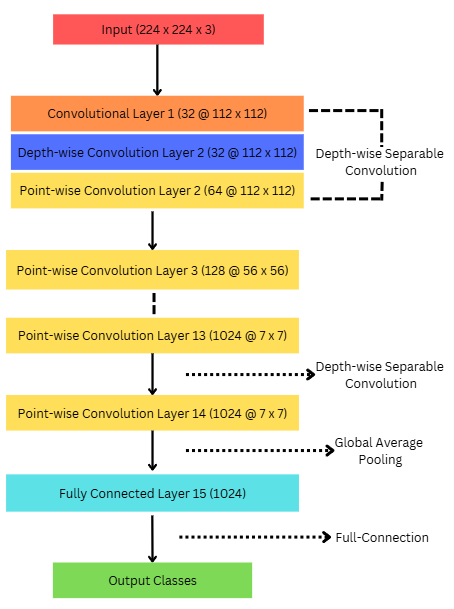}\\
  \caption{Architecture of MobileNet Model, based on depth-wise separable filters}\label{graph}
  \end{center}
\end{figure}

The varying architectures provided valuable insights into the trade-offs between model complexity, and overall performance, enabling the selection of the most suitable model for the OCR task at hand.

\subsection{User-Interface Design}
The User Interface (UI) was meticulously crafted with a user-centered design philosophy, emphasizing clarity, accessibility, and engagement while leveraging cutting-edge technology for preserving India's epigraphical heritage. Its tools and features were designed to seamlessly blend form and function, providing users with intuitive and efficient interactions \cite{UI_1}. 

The three key tool developed include Abhiñāna, an OCR tool designed specifically for extracting text from Ashokan Brahmi inscriptions; Akṣarāntara, a script transliteration tool that enables conversion between ancient scripts and modern languages; and Kośa, a comprehensive database serving as a repository for India's epigraphical records. These tools together facilitate the study and preservation of rich epigraphical heritage.

The clean, minimalistic design, with a dark color scheme enhances readability and focuses attention on the content. The design is highly responsive, adapting to various devices and screen sizes to provide a seamless experience, whether on a desktop, tablet, or smartphone. This ensures that users can access the tools and information wherever they are, supporting the mission to make epigraphical heritage widely available \cite{UI_2}.

The layout is instinctive, with clear navigation and logical content organization. Each page is designed with a specific purpose, guiding users through the website in a way that minimizes cognitive load and enhances the user experience. The transliteration tool page is simple yet functional, providing users with a straightforward interface to convert text between scripts efficiently.

The design prioritizes user needs, offering easy-to-use tools that cater to both scholars and casual users interested in epigraphy. The incorporation of a 'Scan Photo' feature allows users to interact with the OCR tool directly from their devices, while the 'Upload Photo' option caters to those with existing images.



Figure 2 shows the entire methodology of the study.

\begin{figure}[H]
  \begin{center}
  \includegraphics[width=4.8in]{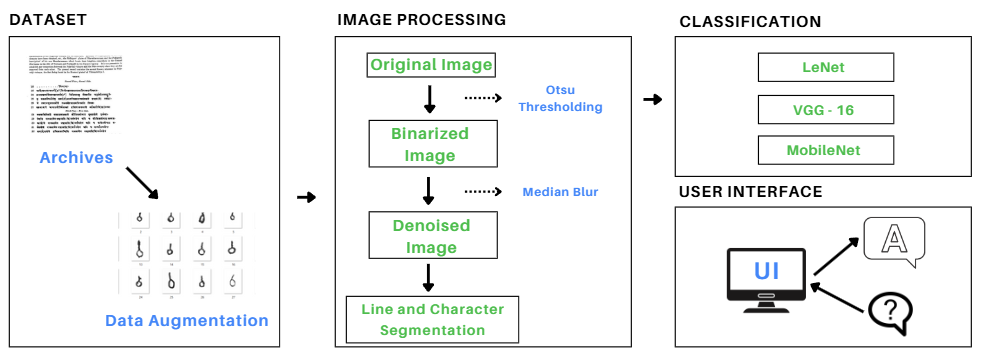}\\
  \caption{Flowchart showing the entire methodology used.}\label{graph}
  \end{center}
\end{figure}

\section{Results}

\subsection{Data Augmentation}
Figure 3 illustrates an example of various augmentations applied to the Brahmi letter "Bha". It includes rotation, shifting, and scaling, which introduce variations in the dataset, aimed at enhancing the robustness of the model during training.

\begin{figure}[H]
  \begin{center}
  \includegraphics[width=3.5in]{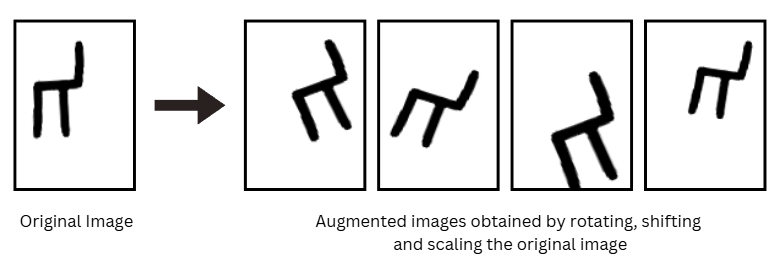}\\
  \caption{Original and augmented images of Brahmi letter Bha.}\label{graph}
  \end{center}
\end{figure}

\subsection{Image Preprocessing}
An estampage of an Ashokan inscription in its original Brahmi script is shown in figure 4. Figure 5 presents the same image after preprocessing, where techniques such as median blur and Otsu thresholding were applied. These methods enhance the contrast and clarity of the inscription, making it more suitable for subsequent analysis. This preprocessing step improves the accuracy of character recognition by reducing background noise and highlighting the script's features.

\begin{figure}[H]
  \begin{center}
  \includegraphics[width=3in]{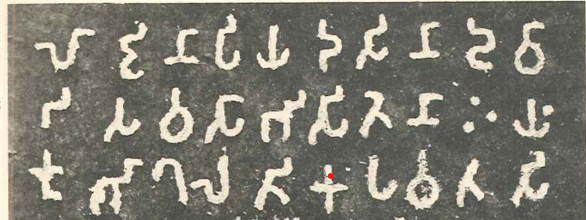}\\
  \caption{Image of an estampage of an Ashoka inscription from the Barabar Caves. }\label{graph}
  \end{center}
\end{figure}

\begin{figure}[H]
  \begin{center}
  \includegraphics[width=3in]{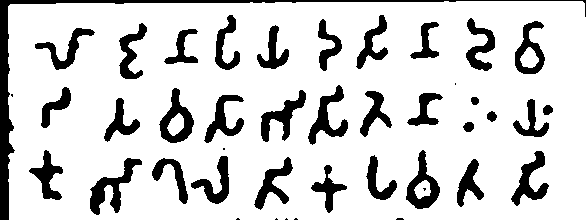}\\
  \caption{Image after removing noise and thresholding.}\label{graph}
  \end{center}
\end{figure}

\subsection{Image Segmentation}
Figure 6 shows the results of image segmentation using both horizontal and vertical projection profile method. This technique is commonly used in optical character recognition to isolate individual characters for further processing.

\begin{figure}[H]
  \begin{center}
  \includegraphics[width=3.5in]{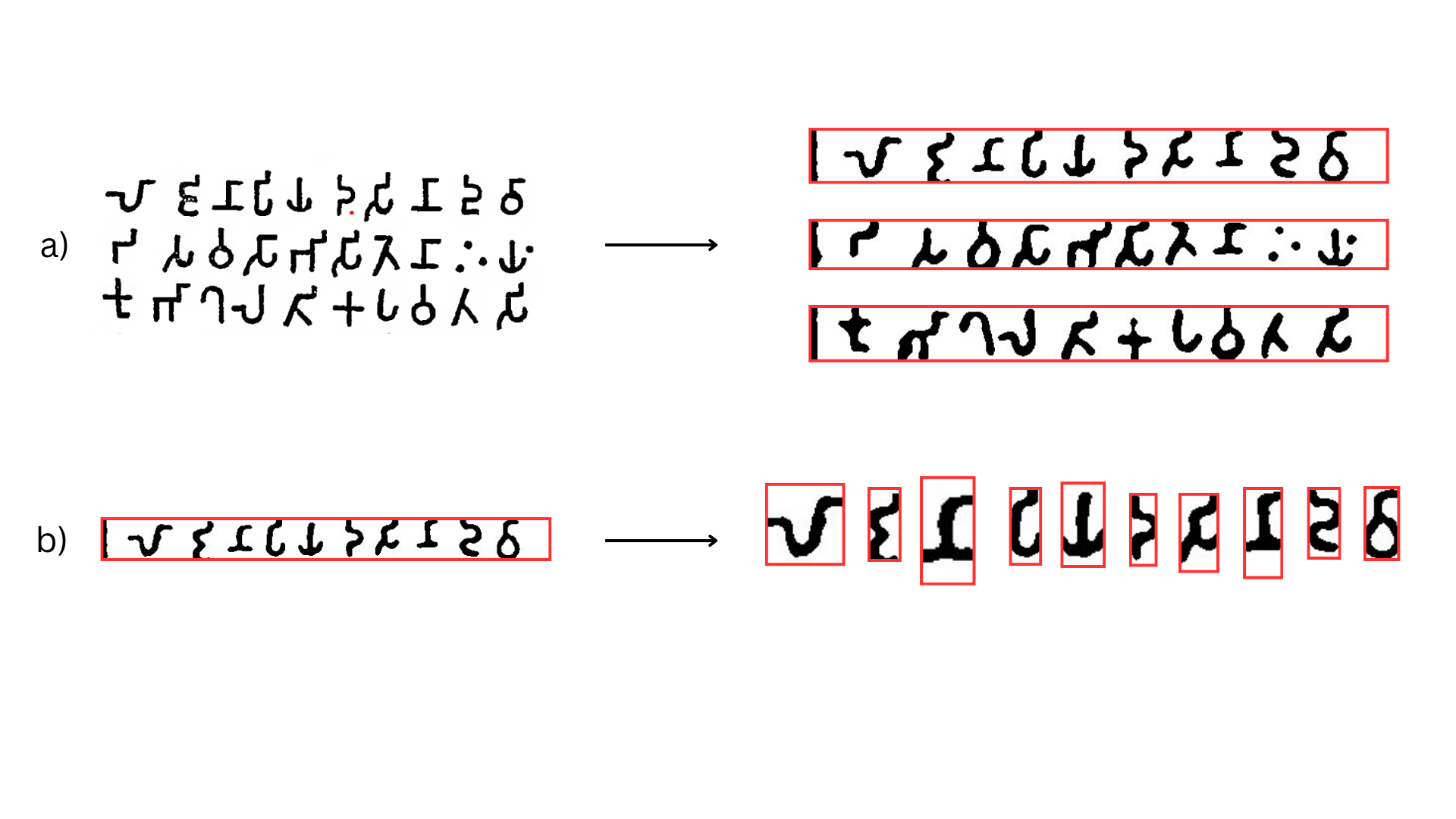}\\
  \caption{Segmentation a) Horizontal projection profile for line segmentation b) Vertical projection profile for character segmentation}\label{graph}
  \end{center}
\end{figure}

\subsection{Classification Models}

The summary the validation accuracies and losses observed for 3 different deep learning models is shown in table 1. The result suggests that MobileNet with average pooling is the most effective model for the given dataset and task. The graph in figure 7 shows the training and validation accuracy of the MobileNet model over 50 epochs. Figure 8 graph shows the training and validation loss of the MobileNet model over 50 epochs. Both training and validation losses decrease significantly, indicating that the model is learning effectively.

\begin{table}[H]
\centering
\caption{Validation accuracy and loss of models used.} \label{table}
\begin{tabular}{|c|c|c|}
\hline
\textbf{Model} & \textbf{Validation Accuracy} & \textbf{Validation Loss}\\
\hline
LeNet & 70.89\% & 1.0673 \\ \hline
VGG-16 & 65.82\% & 1.2654 \\ \hline
MobileNet (with max pooling) & 92.43\% & 0.2507 \\ \hline
MobileNet (with average pooling) & 95.94\% & 0.129 \\
\hline
\end{tabular}
\label{tab:results}
\end{table}

\begin{figure}[H]
  \begin{center}
  \includegraphics[width=3.5in]{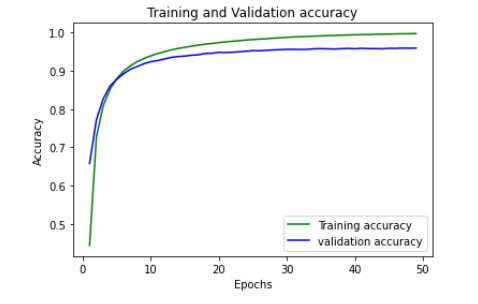}\\
  \caption{Training and validation accuracy of the MobileNet model}\label{graph}
  \end{center}
\end{figure}

\begin{figure}[H]
  \begin{center}
  \includegraphics[width=3.5in]{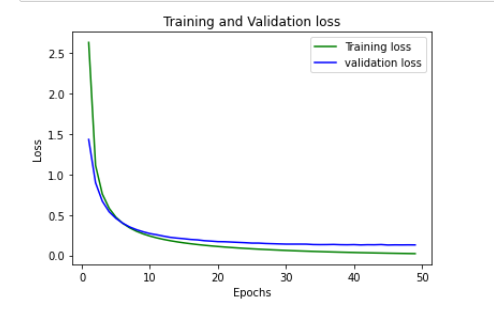}\\
  \caption{Training and validation loss of the MobileNet model}\label{graph}
  \end{center}
\end{figure}

\subsection{User Interface}
Figure 9 showcases the clean and intuitive homepage of the developed user interface. The navigation bar at the top provides easy access to different sections, while the prominent header and tagline highlight the project's mission of preserving, protecting, and promoting epigraphical heritage. The background image, featuring ancient inscriptions, further reinforces the website's theme and purpose. 

Figure 10 depicts a user-friendly interface of the Abhiñāna tool with a clear and concise layout. The prominent "Scan Photo" and "Upload Photo" buttons encourage user interaction and make it easy to access the functionality. 


\begin{figure}[H]
  \begin{center}
  \includegraphics[width=4in]{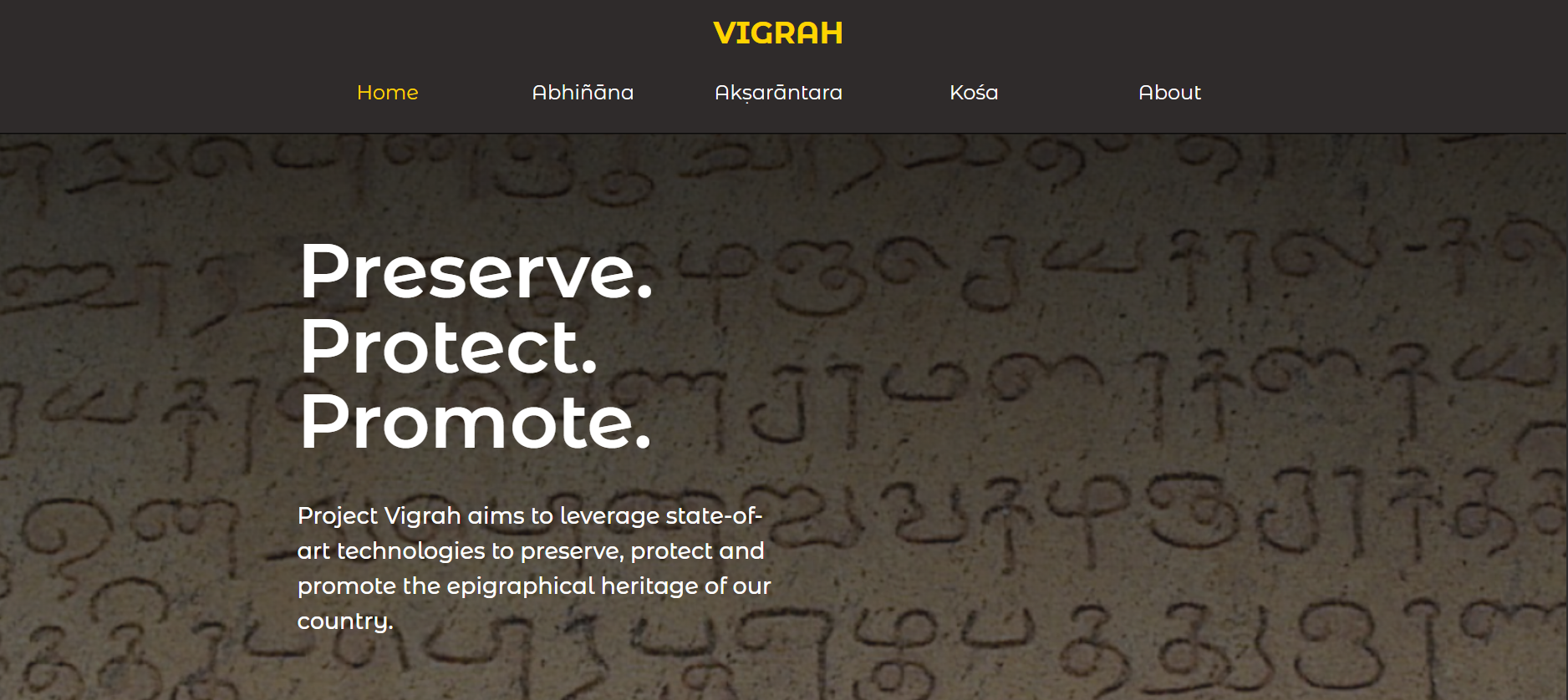}\\
  \caption{The home page of the developed user interface.}\label{graph}
  \end{center}
\end{figure}

\begin{figure}[H]
  \begin{center}
  \includegraphics[width=4in]{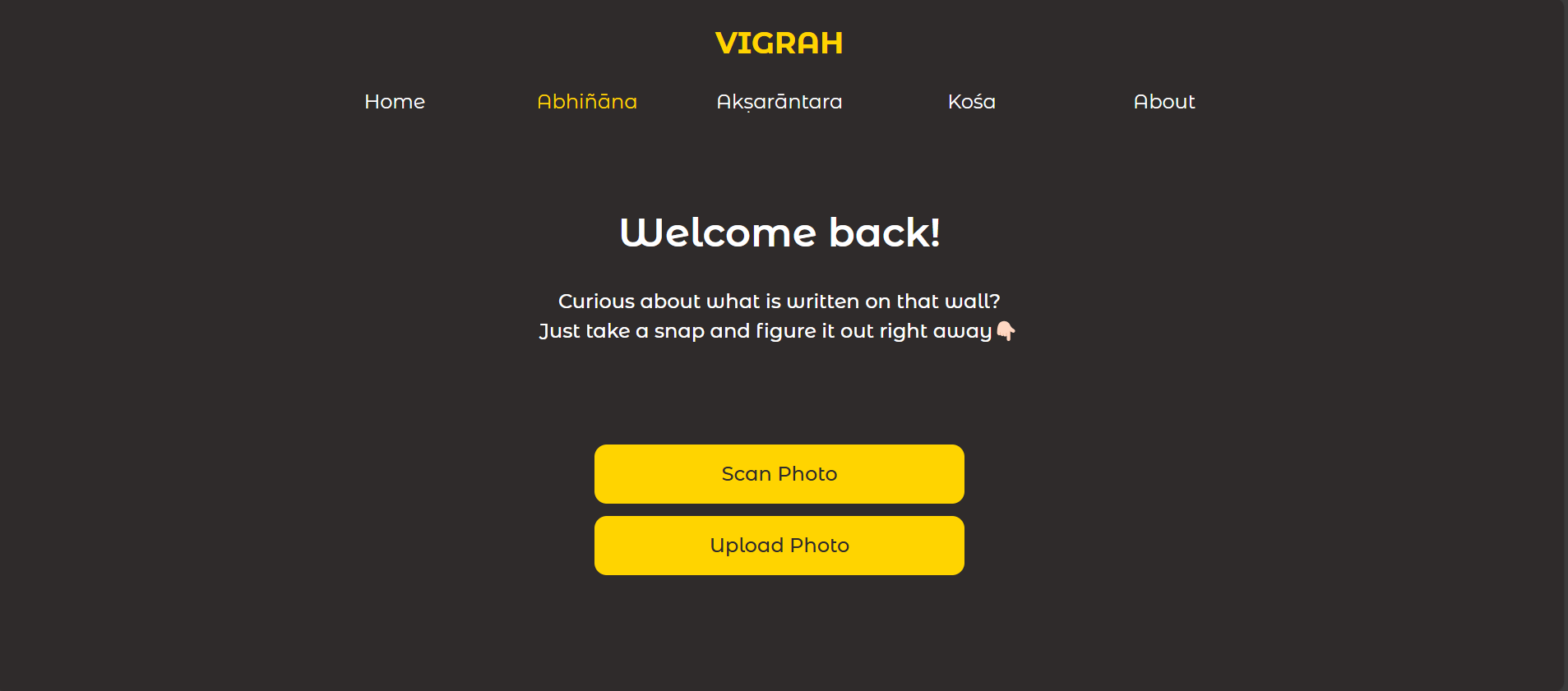}\\
  \caption{The website page showcasing Abhinana tool.}\label{graph}
  \end{center}
\end{figure}
 
\section{Conclusion}
After conducting a comprehensive analysis of Convolutional Neural Network architectures, including LeNet, VGG-16, and MobileNet, it was determined that the MobileNet model trained with average pooling outperformed the others, achieving the highest accuracy metrics on the given dataset. This result highlights MobileNet's efficiency, particularly in resource-constrained environments, making it an ideal choice for mobile and embedded applications.

Future work will focus on refining the OCR system by integrating more advanced deep learning techniques and expanding the dataset to include a broader range of ancient scripts. Additionally, exploring the potential of hybrid models and incorporating attention mechanisms could further enhance the system's accuracy and adaptability across diverse epigraphical contexts, ultimately contributing to the preservation of a wider array of historical inscriptions.

\subsubsection{Acknowledgements} The Indoskript - Manuscripts website  was a valuable resource for this research as it provided a wealth of information about the manuscripts and the characters they contained. The opportunity to access and use the Ashokan Brahmi characters from this website for the research was greatly appreciated.

%
%
%

\end{document}